\documentclass[conference]{IEEEtran}
\IEEEoverridecommandlockouts
\usepackage{cite}
\usepackage{amsmath,amssymb,amsfonts}
\usepackage{algorithmic}
\usepackage{graphicx}
\usepackage{textcomp}
\usepackage{xcolor}
\def\BibTeX{{\rm B\kern-.05em{\sc i\kern-.025em b}\kern-.08em
    T\kern-.1667em\lower.7ex\hbox{E}\kern-.125emX}}
    
\usepackage{authblk}
\usepackage{fancyhdr}
\begin{document}

\title{Predicting Driver Self-Reported Stress by \\Analyzing the Road Scene
}

\author[1]{Cristina Bustos}
\author[2]{Neska Elhaouij}
\author[1]{Albert Sol\'e-Ribalta}
\author[1]{\\Javier Borge-Holthoefer}
\author[1]{Agata Lapedriza}
\author[2]{Rosalind Picard}
\affil[1]{Universitat Oberta de Catalunya \authorcr \{mbustos,asolerib,jborgeh,alapedriza\}@uoc.edu}
\affil[2]{Media Lab, MIT \authorcr \{neska, picard\}@media.mit.edu}

\maketitle
\thispagestyle{fancy}

\begin{abstract}
Several studies have shown the relevance of biosignals in driver stress recognition.  In this work, we examine something important that has been less frequently explored: We develop methods to test if the visual driving scene can be used to estimate a drivers' subjective stress levels. For this purpose, we use the AffectiveROAD video recordings and their corresponding stress labels, a continuous human-driver-provided stress metric. We use the common class discretization for stress, dividing its continuous values into three classes: low, medium, and high. 
We design and evaluate three computer vision modeling approaches to classify the driver's stress levels:
(1) object presence features, where features are computed using automatic scene segmentation; (2) end-to-end image classification; and (3) end-to-end video classification. All three approaches show promising results, suggesting that it is possible to approximate the drivers' subjective stress from the information found in the visual scene. We observe that the video classification, which processes the temporal information integrated with the visual information, obtains the highest accuracy of $0.72$, compared to a random baseline accuracy of $0.33$ when tested on a set of nine drivers.

\end{abstract}


\section{Introduction}

Understanding how the driving scene impacts the driver's emotional state has found a growing interest in the field of driver-assistance technologies.
External driving conditions influence the driver's affective state \cite{Lyu2017,laumann2003selective,mehler2009impact,
jeon2014effects,hennessy1999traffic}, impacting both road safety
\cite{jeon2014effects,Jeon2016} and driver experience \cite{Magana2020}. While some specific case studies show that the drivers' stress correlates to road traffic conditions \cite{hennessy1997relationship, hill2007driver,Bitkina2019, chung2019methods} and road type (city, highway, and parking) \cite{Healey2005,Liu2018,
ElHaouij2018,Bitkina2019}, automatically inferring these, possibly causal, relations directly from the driving scene has not yet been explored in-depth.

In this paper, we study how the drivers' subjective stress level can be estimated from images displaying the driving scene. Our study takes inspiration from Healey and Picard's original study on driver stress \cite{Healey2005}, which constructed low, medium and high stress conditions corresponding respectively to sitting in a parked car with eyes closed, driving on a highway under optimal conditions (dry pavement, no traffic or construction, and good weather), and city driving in a busy area. The authors also measured the complexity of events minute-by-minute under each condition (e.g. a turn, a pothole, or a pedestrian are example events that increased the complexity).  Measurements of stress-related physiology and self-reported stress agreed with the low, medium, and high-stress levels as well as correlated with the human-rated assessment of the scene complexity. Based on these observations and recent research \cite{ElhaouijPhDThesis}, our hypothesis is that some objects and events that are visible in the driving scene are informative enough to approximate the current drivers' subjective stress level.

More concretely, the goal is to empirically test with different supervised machine learning approaches, whether the drivers' subjective stress can be inferred solely from real driving scenes extracted from the AffectiveROAD dataset \cite{WebsiteAffectiveRoad}. The dataset contains a collection of real driving videos taken by a camera frontally pointing to the road labelled with the drivers ``stress'' signals. 

\begin{figure}[htbp]
\centerline{\includegraphics[width=0.5\textwidth]{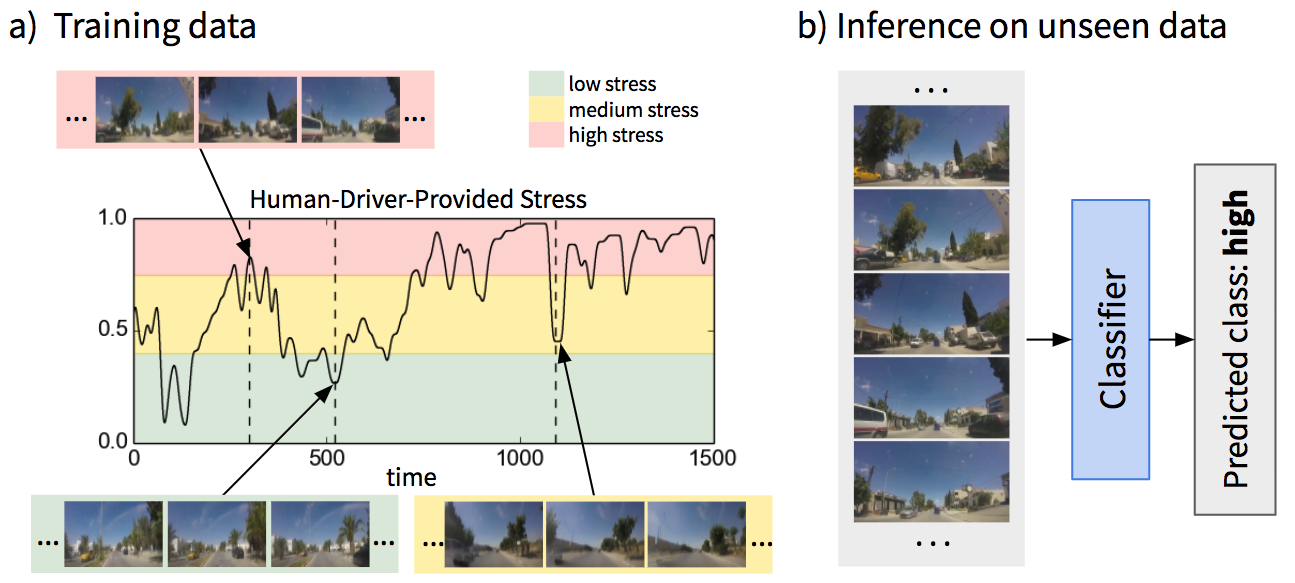}}
\caption{(a) Illustration of the data used in our study and (b) the prediction of drivers' subjective stress level on new unseen driving scene videos.}
\label{fig:pipeline}
\end{figure}

The provided ``stress'' signals were constructed in real-time by an observer who sat in the rear seat and annotated the driving scene ``complexity'', and validated post-experience by the driver \cite{ElHaouij2018}. 
Fig.~\ref{fig:pipeline}.a illustrates the data used in our study, while Fig.~\ref{fig:pipeline}.b illustrates the inference of drivers' subjective stress level on new unseen driving scene videos.

The three modelling approaches differ in complexity and technique. First, we used classical machine learning (Random Forest and Support Vector Machines) on $66$ handcrafted features encoding the presence of common objects, such as cars, road, traffic signals, or pedestrians. 
The second and third modelling approaches are based on two end-to-end Deep Convolutional Networks: an image Convolutional Neural Network (CNN) and a video CNN. 

We examine the three modelling approaches and we find all perform significantly above chance on the tested task. As expected, the best accuracies are obtained by the video CNN. 
For the video CNN we also use a CNN explainability technique (GradCam \cite{selvaraju2017grad}, described in Sect.\ref{sec:methodology_cam}) to visualize the areas of the input frames that contribute the most to the output of the model. We find that the model puts attention on certain objects and specific configurations of the scene that align with our work hypothesis. 

Our study shows promising results on using the driver scene to approximate drivers'  stress. Such findings may have practical applications: from intelligent cars able to better assist drivers under stress, to suggesting less stressful routes and helping improve driver safety.

\section{Related Work}
\label{sec:realted_work}

Detection of driver affective state is relevant for the development of in-vehicle systems that can help improve driving experience \cite{ZepfSurvey}. Stress can have a significant negative impact on driving performance, causing traffic violations and crashes \cite{Simon1996,Bowen2020}. Various scenarios and interventions have been designed to alleviate the ``extreme affective states'' when detected \cite{Nass2005,Hernandez2014,Paredes2018}. 
Based on the driver's state taxonomy of Braun et al. \cite{Braun2019}, 
these extreme states correspond to dangerous states, while states with medium arousal levels and positive valence are recognized as optimal ones.

Different approaches have been used to detect driver's stress and affective states, including physiology, facial expression, self-reports or biosignals \cite{Rastgoo2018,
ZepfSurvey,Nvemcova2020}, but most have focused only on sensing momentary changing signals from the driver. Including contextual parameters, whether internal or environmental \cite{Rastgoo2018}, is expected to help improve accuracy. Internal context parameters may include personal parameters such as driver mood or personality \cite{Jeon2016,Magana2020},
while environmental parameters typically characterize external factors such as the weather
 \cite{hill2007driver,Rimini2001}, road traffic \cite{hennessy1997relationship, hill2007driver,Bitkina2019, chung2019methods}, and road type \cite{Healey2005,Liu2018,ElHaouij2018,Bitkina2019}.  Our work in this paper focuses on vision-based extraction of environmental context. 

Urban settings tend to have higher complexity \cite{bustos2021explainable} that may require a higher level of attention which, in turn, requires higher cognitive workload, thus usually increasing driver arousal and stress. 
Features that represent this complexity, directly extracted from the visual scene, may be used to characterize such external conditions. Thus, tools for scene analysis become increasingly important to assess the driver's affective state. Thanks to recent advances in machine learning \cite{Ren2016,Zhao2017} with shared real-world datasets \cite{Geiger2012,zhou2017places,Ding2020}, we believe it is now feasible to train an automated system to predict a driving-induced state of stress from a visual scene.

The definition and annotation of driver emotion remains challenging in real-world driving settings. According to a recent survey by Zepf et al. \cite{ZepfSurvey}, reliable annotations are important to effectively recognize the emotional state of drivers. The authors identified three main approaches to such annotations based on their survey: self-reports, external annotators, and experimental context. Self reports require the involvement of the participants by usually asking them to report their driving experience (after the drive) according to questionnaires such as: Positive and Negative Affect Scale \cite{Kato2011} or Self-Assessment Manikin \cite{Ihme2018}. This approach is subjective and might induce some biases. The annotation approach based on external annotators can be more reliable;  however, it requires time, effort, and extra cost to find, train, and compensate experienced observers. 

If it were possible to automatically identify experimental contexts that are reliably associated with stress levels, then it could provide a new method to annotate a driver's likely state.  Thus, our work may help not only with predicting driver stress in real-time applications, but also it may help in expanding the utility of other unlabeled data sets for additional research. 

\section{Data}
\label{sec:data}

The research we develop in this paper is based on the AffectiveROAD dataset \cite{WebsiteAffectiveRoad} which includes gold-standard stress annotations provided by drivers after real-world open-road driving experiences. We describe below the acquisition protocol and the details of the dataset. 

\subsection{AffectiveROAD data description}
The AffectiveROAD protocol aims to jointly collect, in real driving scenarios, information about the driving environment (in and outside the car), driver's physiological state (for example, electrodermal activity, breathing rate and heart rate) and the driver's contextual ``stress'' levels.

\begin{figure}[htbp]
\centerline{\includegraphics[width=0.49\textwidth]{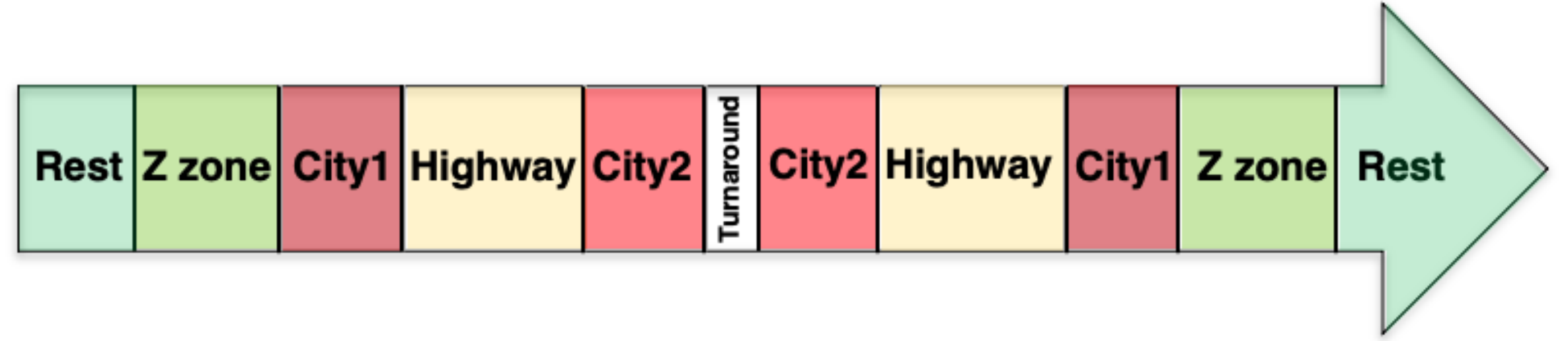}}
\caption{Route proposed to the AffectiveROAD study participants.}
\label{fig:Roadmap}
\end{figure}

The protocol is designed to obtain information about typical daily trips in Grand Tunis (with normal traffic conditions) of $31$ km in length that take about $85$ minutes (including $30$ minutes  rest). Driving routes (see Fig. \ref{fig:Roadmap}) are chosen so that drivers alternate between different road types and environments, which may induce different stress levels. 

\begin{figure}[htbp]
\centerline{\includegraphics[width=0.51\textwidth]{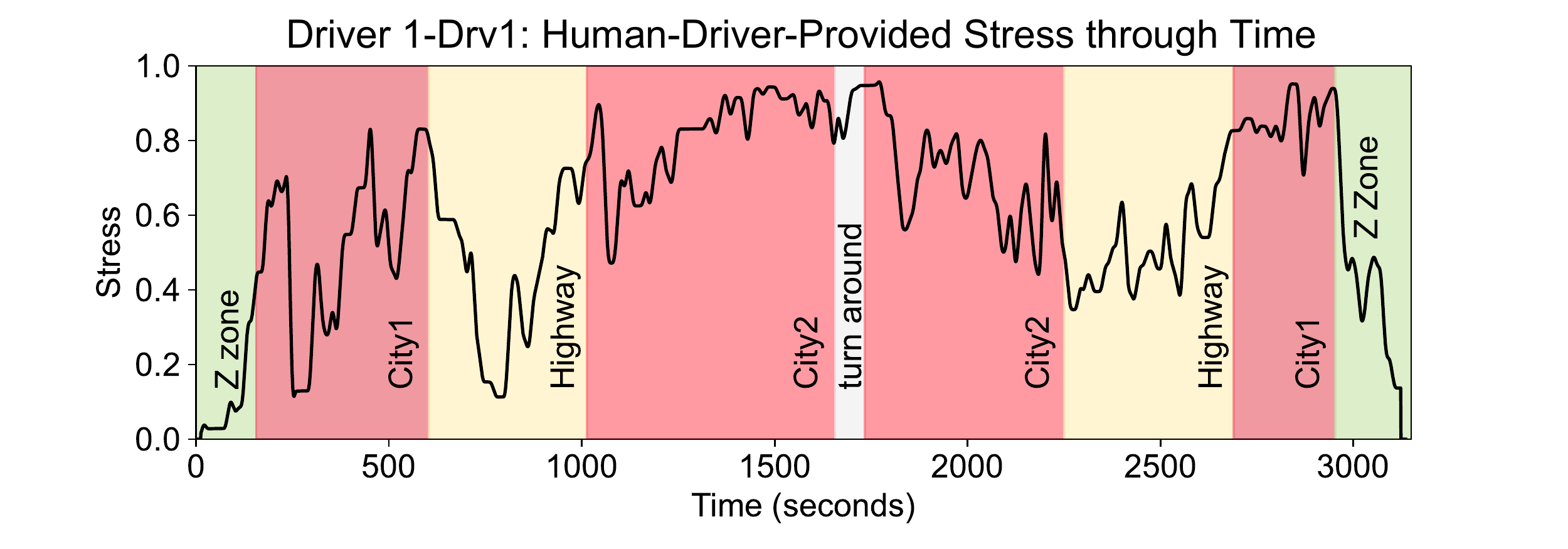}}
\caption{Example of human-driver-provided stress measure for driver 1. Background color in the plot indicates the part of the road map.}
\label{fig:example_self_report}
\end{figure}

During each drive, an experimenter, who sat in the back seat, annotated in real-time the perceived stress using a laptop-based slider. This resulted in a subjective score\cite{ElHaouij2018}, ranging from $0$ (not stressful) to $1$ (extremely stressful), based on the perception of the driving scene ``complexity'' and the stress level of the driver. See Fig. \ref{fig:example_self_report} for an example of the evolution of the subjective score during the driving experiment. The drivers were asked to validate or correct 
the score after the experiment. Synchronized videos captured both the inside and outside car environments (visualized side by side) and these were synchronized with the stress score and were shown to each driver. These videos and stress scores were embedded in a platform that offered the user the option to change the stress metric values at any point of the experiment. The resulting stress metric will be denoted as the "human-driver-provided stress" in this paper.  

The complete dataset provides data for $13$ paths completed by 9 drivers in sunny days. Six of the paths were performed by drivers who completed the driving trajectory only once, and seven paths were accomplished by three drivers who repeated the experiments on different days. The complete dataset related to the $13$ paths contains $676.3$ minutes of video recordings.

\subsection{Data subset used in this study}

Although the AffectiveROAD dataset contains various modalities, the work in this paper uses only the human-driver-provided stress metric and the corresponding scene road video recordings, because it provides cleaner labels on the driver’s stress than the physiological signals. 

The stress metric was obtained as follows.

\begin{figure}[htbp]
\centerline{\includegraphics[width=0.5\textwidth]{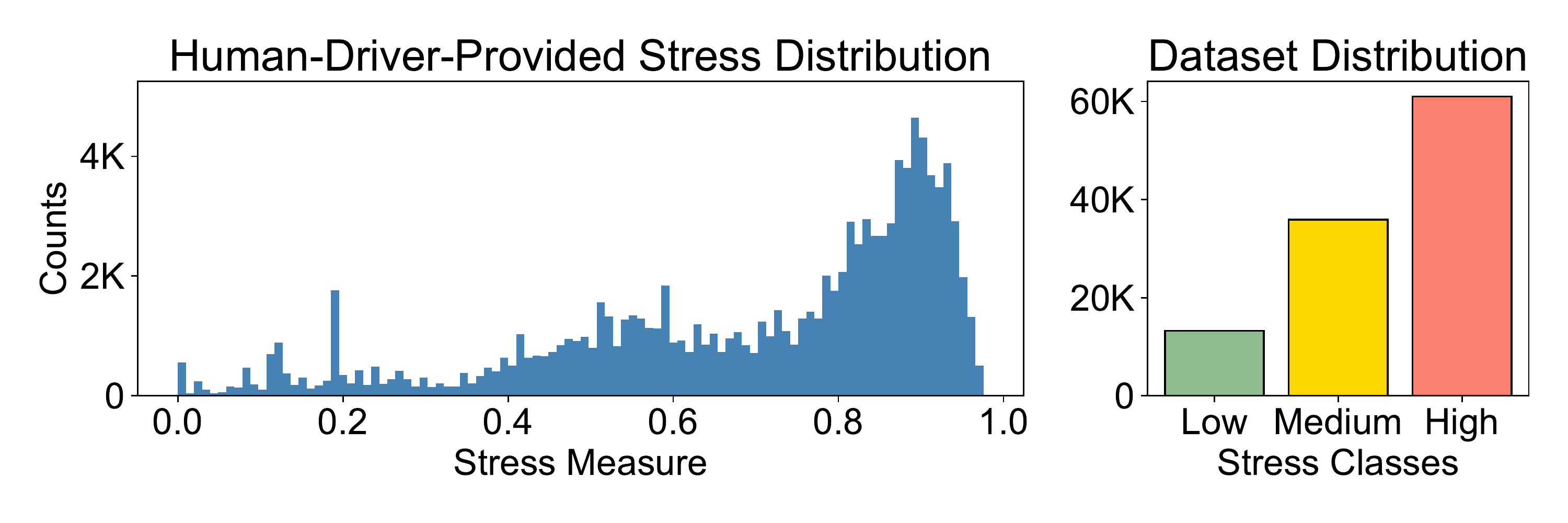}}
\caption{Dataset Distribution. Left: frequency distribution histogram of continuous human-driver-provided stress. Right: histogram of discretized human-driver-provided stress.}
\label{fig:data_distribution}
\end{figure}

In Fig. \ref{fig:data_distribution}, the left plot shows the histogram of the original stress measures provided by the drivers. Note that the stress variability of the original measures was high for all the drives. It ranges between $0$ and almost $1$. We then min-max normalized the stress metric within each driver, so each person's labels ranged from a minimum of $0$ to a maximum of $1$ (original min and max value were very close to 0 and 1, respectively, for all the drivers).  
Based on the 3-modal metric distribution shown in Fig \ref{fig:data_distribution} and after consulting the experimenter who generated the continuous metric \cite{ElHaouij2018}, we constructed three discrete stress classes as follows: {\bf low} stress class, for stress scores between $0$ and $0.4$; {\bf medium} stress class, for scores between $0.4$ and $0.75$; and {\bf high} stress class for scores higher than $0.75$. In Fig. \ref{fig:data_distribution}, the right plot shows the number of instances per each of the three stress classes.

\begin{figure}[htbp]
\centerline{\includegraphics[width=0.5\textwidth]{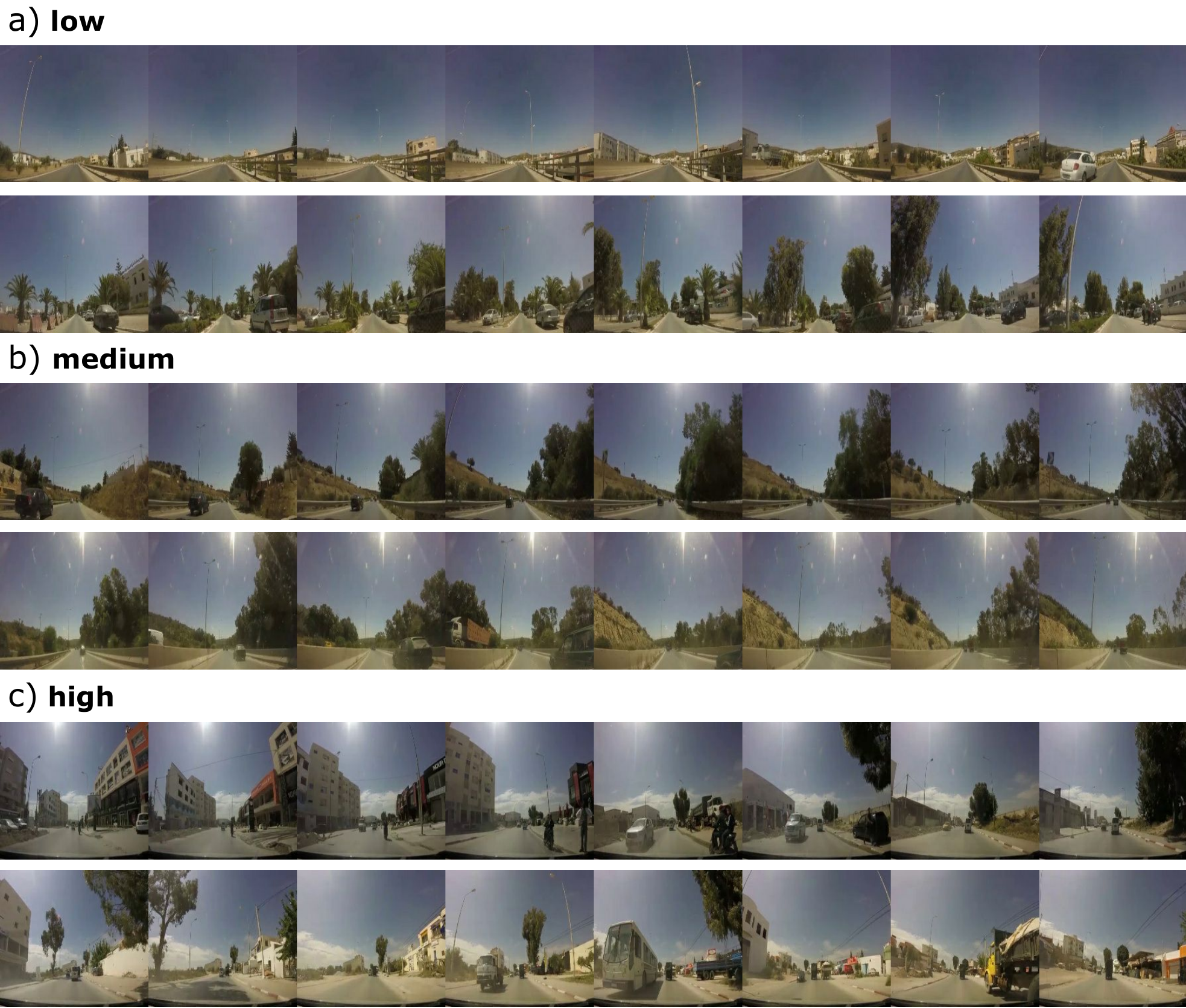}}
\caption{Examples of frame sequences for low (a), medium (b), and high (c) stress measure classes.}
\label{fig:examples_classes}
\end{figure}

For illustrative purposes, Fig.~\ref{fig:examples_classes} shows randomly selected examples of video sequences for the defined low, medium, and high stress categories. Notice that we can already observe some visual differences in these examples. For the low stress category we do not see much traffic, and the video sequences correspond to scenes where the speed is slow. Based on this extracted sample of video sequences, the low stress category includes segments from the Z zone and highway as defined in  Fig. \ref{fig:Roadmap}. For the medium stress category, we observe the road scene corresponding to areas where one can circulate at a higher speed without close vehicles. The frame sequences depicted in the medium section of Fig.~\ref{fig:examples_classes} correspond mainly to highway segments. Finally, for the high stress category we observe more clutter, buildings,s traffic, and pedestrians. Our expectation is that these types of characteristic patterns that we observe for the different stress measure categories can be captured by a computer vision model attempting to categorize road scene images or video sequences into the corresponding low, medium, or high stress categories.

To avoid the habituation effect that could be induced due to the repetition of the experiment and that might affect the perceived stress, in our experiments we only considered the video recordings corresponding to the first drive of each participant. Thus only these first 9 drives were selected and they correspond to the following participants codes: 1.Drv1-1, 2.Drv2-1, 3.Drv3-1, 4.Drv4-1, 5.Drv5-1, 7.Drv6-1, 9.Drv7-1, 10.Drv8-1, and 11.Drv9-1. 
For video modelling, the original video sequences, taken at 25fps, were reduced to 2fps.

\section{Methodology}
\label{sec:methodology}

This section describes the different methods we use throughout the paper. First we provide details about the data splits used for the training, validation and test sets. Then, we explain how we perform automatic object segmentation of road scene images. Later, we describe the three different approaches we use for the modelling. Finally, we explain the interpretability techniques we use to qualitatively relate urban image patterns and classification scores. 

\subsection{Data Splits}

All modelling experiments consider a total of 9 data splits, each leaving out one driver as later testing data. Then, among the remaining 8 drivers, the videos of 2 randomly selected drivers are used for the validation set, while the videos of the remaining 6 drivers are used to train. Each of these $9$ data splits is denoted by $D_i$, for indexes $i$ corresponding to the experiment ID in the original AffectiveROAD dataset. Notice that with this data-split protocol we evaluate to what extent the modeling of the human-driver-provided stress measure generalizes to unseen drivers. For reproducibility, the validation and testing drivers for each data split are provided in Tab.~\ref{tab:data_split} (at each data split the remaining 6 drivers are used for training).


\begin{table}
	\caption{Data split sets using for the different training experiments}
	\centering
	\begin{tabular}{|c|c|c|c|}
		\hline
	     {\bf Test ID} & Validation Drivers & Testing Drivers \\
		\hline \hline
		$D_1$ &  2.Drv2-1, 10.Drv8-1  &  1.Drv1-1   \\
        \hline
		$D_2$ &  3.Drv3-1, 11.Drv9-1  &  2.Drv2-1  \\
		\hline
		$D_3$ & 1.Drv1-1, 9.Drv7-1 & 3.Drv3-1 \\
		\hline
		$D_4$ & 9.Drv7-1, 2.Drv2-1  &  4.Drv4-1 \\
		\hline
		$D_5$ & 1.Drv1-1, 11.Drv9-1 & 5.Drv5-1 \\
		\hline
		$D_7$ & 4.Drv4-1, 10.Drv8-1  &  7.Drv6-1 \\
		\hline
		$D_9$ & 3.Drv3-1, 5.Drv5-1  &  9.Drv7-1 \\
		\hline
		$D_{10}$ & 7.Drv6-1, 5.Drv5-1  &  10.Drv8-1 \\
		\hline
		$D_{11}$ & 3.Drv3-1, 4.Drv4-1 & 11.Drv9-1 \\
		\hline
	\end{tabular}
	\label{tab:data_split}
\end{table}


The distribution of low, medium, and high stress classes in each of the data splits is, approximately, $12\%$, $33\%$, and $55\%$, respectively. The dataset has a total of almost 110K instances. In our experiments we balanced the three classes by upsampling the training dataset, each class to the class with more examples, and downsampling the validation and test dataset each class to the class with less examples.  The resulting test sets are thus all balanced so that a random choice classifier would score $0.33$ accuracy on average.


\subsection{Segmentation of urban scenes}
\label{sec:segmentation}

\begin{figure}[htbp]
\centerline{\includegraphics[width=0.5\textwidth]{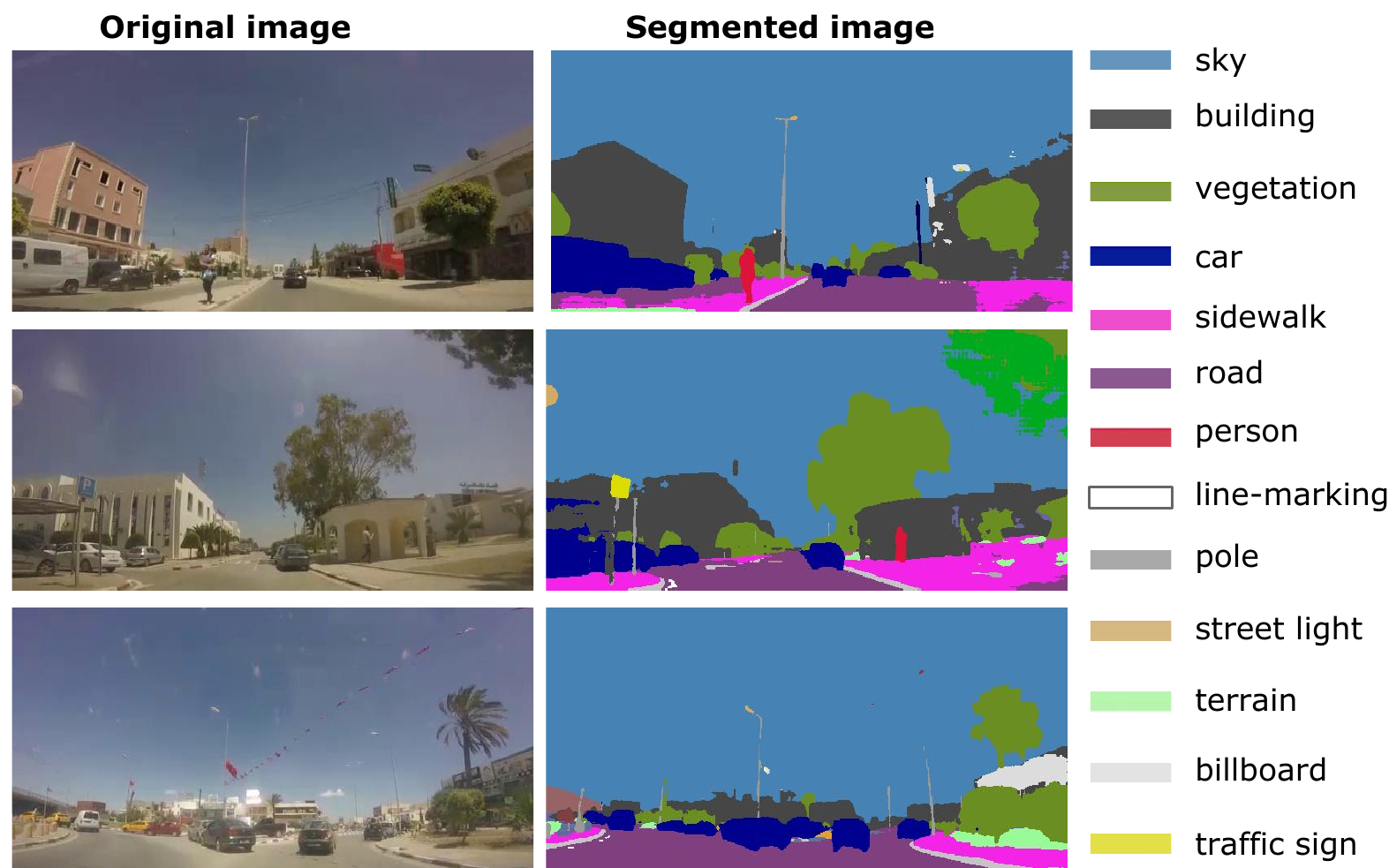}}
\caption{Examples of original images with their segmentation mask.}
\label{fig:examples_segmentations}
\end{figure}

To gain understanding about the impact the road scene has on the human-driver-provided stress, we attempt to identify which objects are present in the visual scene and where they are located. For this goal we require to semantically segment each dataset video frame. Semantic segmentation assigns an object category label to each pixel of an input image. We segmented the images using the Inplace-ABN (DeepLabV3+WideResNet-38) implementation \cite{bulo2018place}, already trained with the Mapillary Vistas dataset \cite{neuhold2017mapillary}. Mapillary Vistas dataset is a large-scale diverse street-view image dataset for semantic urban understanding, containing 25k high resolution images with pixel-accurate annotations of 66 semantic urban objects categories. The Mapillary Vistas dataset has a global coverage and very diverse selection of images considering different weather and seasonal conditions, points of view, and camera models. Annotated objects included categories like sky, road, building, sidewalk, person, and car, among others. Some random examples of images and corresponding segmentations are shown in Fig.~\ref{fig:examples_segmentations}.

\subsection{Modelling Approaches}

To predict driver's stress from their video recordings, we use three different modelling approaches. Two of them take as input a single video frame, and the goal is to estimate the stress measure class for the specific time stamp corresponding to the input frame. The third approach considers instead a video sequence of $n$ seconds as input, and the goal is to estimate the stress measure class at the time stamp corresponding to the last input frame. Thus, in the third case, the model is using the current time stamp frame and some previous frames to infer the stress measure class. For a fair comparison, all modelling approaches were trained with the same number of samples. In the case of the second and the third models, we created a dataset of video sequences with length n = 32 seconds. While the third model uses the video sequences, varying the sequence length from 1 to 32 seconds, the second model uses only the last frame from each sequence. 

\subsubsection{Image classification with object presence features}
\label{sec:meth_object_presence_features}

Our first modelling approach encodes the visual information at each time stamp as a 66-dimension feature vector. Each feature corresponds to one of the object categories included in the segmentation model. More concretely, each feature encodes the area occupied by the corresponding object in the image. With this feature vector we train different classifiers to estimate the stress measure category of the corresponding frame. In particular, we trained Random Forest, Linear SVM, and RBF-SVM using the SciKit-learn Python library. 

\subsubsection{End-to-end image classification}
\label{sec:methodology_image}

Our second approach consists of an end-to-end Convolutional Neural Network (CNN) that takes as input a video frame to infer the human-driver-provided stress measure category corresponding to the same time stamp. In particular, we use a VGG-16 \cite{simonyan2014very}, pretrained with Places-365 dataset \cite{zhou2017places}. Additionally, to add complexity to the model, after the convolutional layers, we added two consecutive fully connected layers (of 512 hidden units) followed each by a dropout layer of 0.5. Lastly, a fully connected layer with 3 hidden units (one per class) was added as a prediction layer, with softmax activation function. At the training phase, which considers the cross-entropy loss, all layers were frozen except the last convolutional block of the VGG-16 and the newly added layers.

\subsubsection{End-to-end video classification}
\label{sec:methodology_video}

\begin{figure}[htbp]
    	\includegraphics[width=0.5\textwidth]{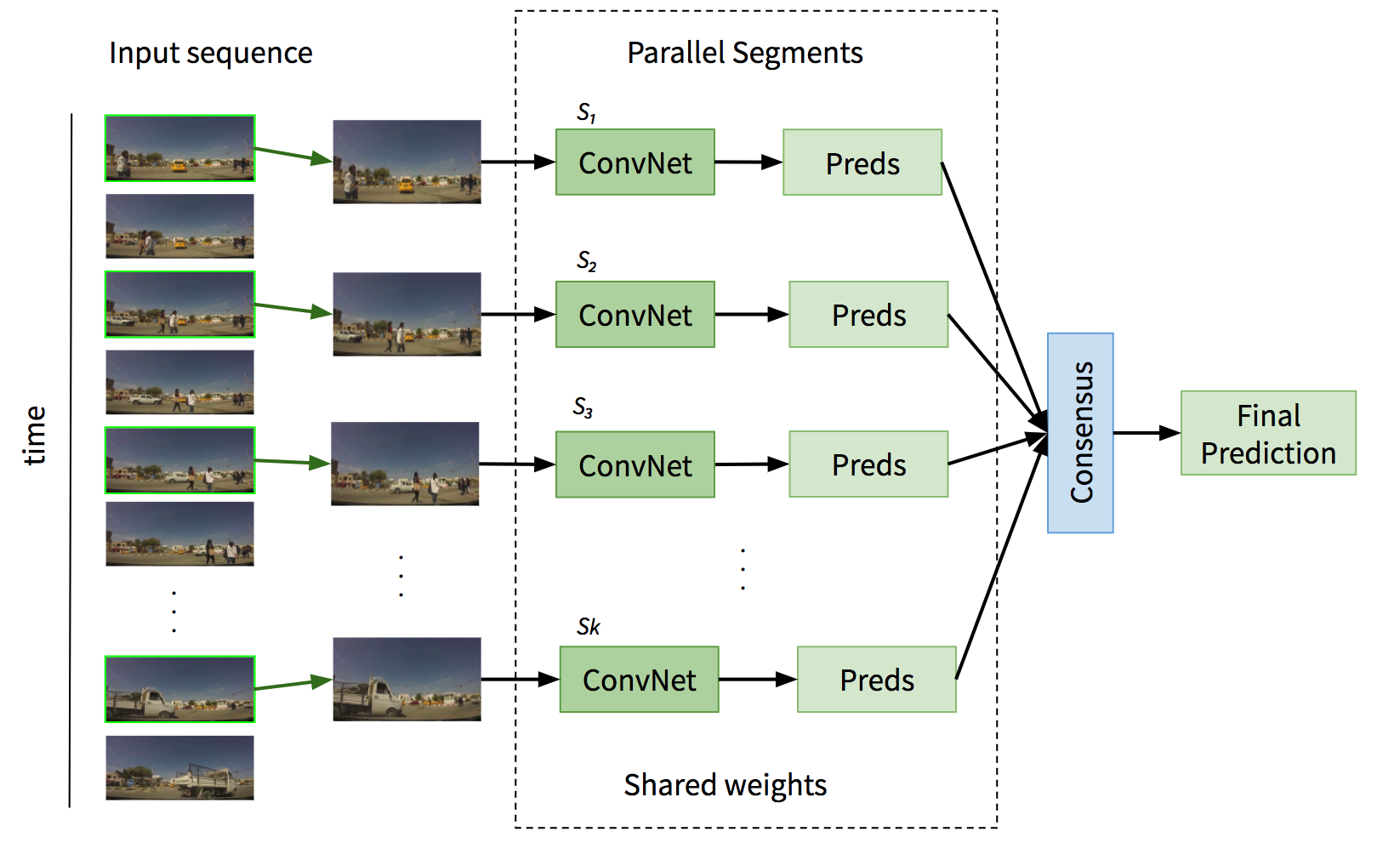}
        \caption{Temporal Segment Network Architecture.}
        \label{fig:tsn}
\end{figure}

Our third approach to model the inference of the human-driver-provided stress measure category is based on Temporal Segment Networks (TSN) \cite{wang2016temporal}. TSN is a video-level framework that was originally proposed for action recognition in videos. TSN aims to model temporal data with segment-based sampling together with an aggregation module called \textit{consensus}. The extra information provided by the video sequence is important to increase the accuracy of classification tasks in several setting, including ours. Formally, applied to our problem, each input sequence (video recordings of a driving), $V$, is divided into $K$ segments of equal length ${S_1,S_2,..,S_k}$. Then, each segment $S_k$ is represented by its first frame, $T_k$. Let $F(T_k,W)$ represent a CNN with parameters $W$, which operates on frame $T_k$ to produce an individual prediction. Let function $G$ represent the aggregation consensus that combines the outputs obtained for each frame and let function $H$ provide the final classification for input sequence $V$. Then the set of all $T_k$ are directly considered by the TSN as follows:
\begin{equation}
TSN(T_1,..,T_k) =  H(G(F(T_1;W),...,F(T_k;W)))
\end{equation}

The implementation of the TSN architecture is shown in Fig. \ref{fig:tsn}. For function $F$ we used the VGG-16 as defined in our image end-to-end model (see Sect.\ref{sec:methodology_image}) and for the consensus module, we used the average. In the training phase, all VGG-16 segments shared the same weights and their training process was equivalent to the one described in Sect.\ref{sec:methodology_image}.

Other implementation details include: we used RMSprop as the optimizer algorithm to learn the network parameters.  The batch size was set to 4 videos and the learning rate parameter was set to 10e-5. All the convolutional networks from the TSN segments were initialized with a VGG16 pretrained with the places-365 dataset. Most of the trainings converged at 10 epochs. For running the experiments, we used a GPU Tesla P100 of 16GB RAM, and 32GB for physical RAM. The approach was programmed in Python, using the Deep Learning framework TensorFlow 2.3.

\subsection{CNN interpretability with Class Activation Maps}
\label{sec:methodology_cam}

Class Activation Mapping (CAM) \cite{zhou2016learning} and related interpretability approaches, such as gradient-weighted CAM (GradCAM \cite{selvaraju2017grad}), are used to visually interpret the output of a CNN. Concretely, CAM heat maps highlight the regions of the image that contributed the most for the classification score for a specific class. While the original formulation of CAM can be just applied to fully convolutional models, GradCAM can be applied even in the presence of fully connected layers after before the output. In this study we use GradCAM to visualize the most informative frame regions for our video CNN Model.

\section{Experiments}
\label{sec:experiments}

This section presents the conducted experiments. First, we study how the urban scene composition relates to the stress measure
and show the results obtained with the three modelling approaches described in Sect. \ref{sec:methodology}. 
Then we present an interpretability analysis using Class Activation Maps.

\begin{figure*}[h]
    	\includegraphics[width=1\textwidth]{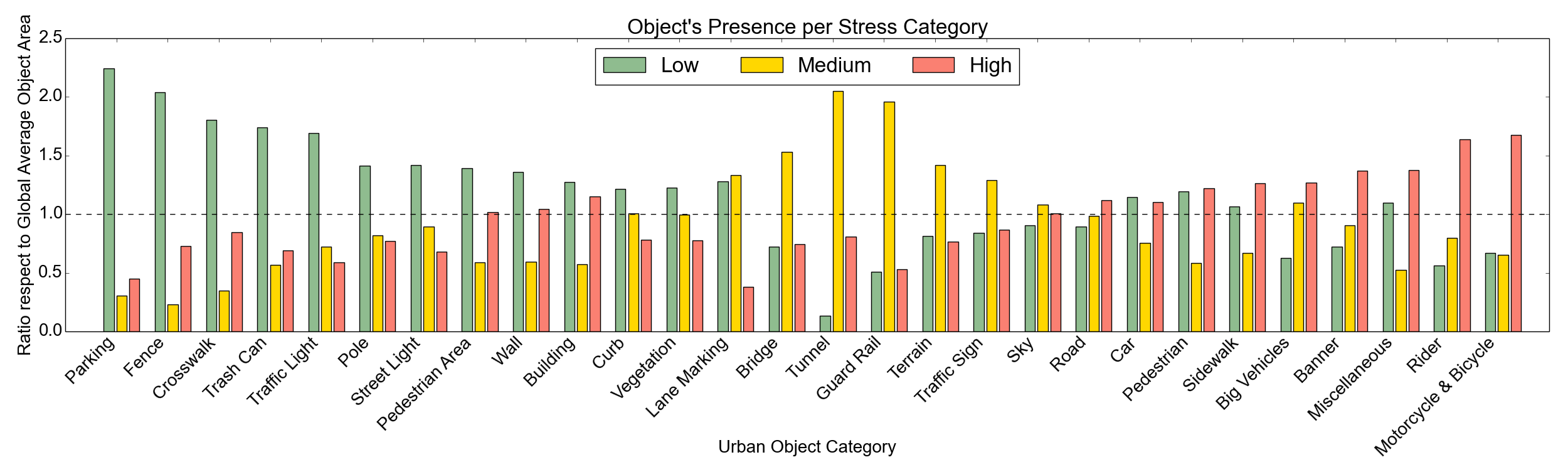}
        \caption{Object's over/under representation level in images tagged with different stress levels. We first obtain the average object occupancy of an object $i$, $\langle o(i) \rangle$, by (1) segmenting each image in our data (see Sect.~\ref{sec:segmentation}), (2) computing the fraction this object represents over the total image and (3) averaging over all images. Then, for each category, $p \in \mbox{\{'low', 'medium', 'high'\}}$, the average occupancy was calculated for each object, $\langle o(i)_p \rangle$.  The plot shows each object's mean on category $p$ with respect to each object's global mean, $\langle o(i)_p \rangle/\langle o(i) \rangle$. Values larger than one indicate the object is over-represented with respect to the global mean and a value lower than one indicates under-representation. Objects have been manually sorted to facilitate interpretability.}

        \label{fig:seg_mean_diff}
\end{figure*}

\subsection{Stress measure and road scene composition}
\label{sec:stress_scene_obj}


We start the experimental section by analyzing which objects are over and under-represented on images related to different stress levels. Considering the automatic image segmentation (as described in Sect.~\ref{sec:methodology}), we compute the ratio between the average presence of each object in low/medium/high stress images with respect to the average presence of the same object across all images. Results are shown in Fig.~\ref{fig:seg_mean_diff}. Notice that when the object ratio is 1 it means that its presence in the given stress class does not differ from the average presence. Ratios larger or smaller than 1 indicate, respectively, more or less object presence than the average. As we can see, each stress condition could be characterized by the over-representation of between 4 to 5 urban objects. 

The parking, fence, crosswalk, and trash can classes are expected to over-represent the low-stress level, since these objects are frequent in the parking lot and Z zone areas. In these settings, the car was either parked before starting the drive, or slowly moving in the Z zone, which is supposed to induce lower stress than more complex settings. 
On the other hand, according to the findings of \cite{bustos2021explainable}, one would expect to find the traffic lights and pedestrian among the urban objects characterizing city segments which are supposed to induce a high-stress level. 
However, for this set of data, both categories over-represent the low-stress level. After a secondary examination of the images corresponding to low-stress level, we noticed two particularities in the AffectiveROAD dataset that explain this observation. First, several pedestrians, sometimes a group of them, are walking around, not necessarily crossing the street, particularly in the Z zone. Second, some of the traffic light poles are short, making them to occupy the majority of the image, leading to an over representation of the traffic light category in the low stress class.

For the medium-stress level, we find those categories that typically define highway driving and related elements, like ramps (entrance or exit road): tunnel, guard rail, bridge, terrain and traffic signs are the main over-represented objects. 

For the high-stress level, we find objects like motorcycle and bicycle, rider, banner, big vehicles and miscellaneous. Riders are the persons on motorcycles and bicycles. This explains why the values of the ratio are almost the same for both rider and motorcycle \& bicycle categories. Banners were present in the city areas. Miscellaneous includes several object categories present in the videos such as bench, ground animal, mailbox, potholes, catch basin, junction box, among others. This means that multiple visible objects and multiple interacting vehicles coexist in the same scene, as found in congested urban environments.

Overall, the representation of objects found per stress category validates the assumptions used in earlier studies that parking, highway, and city driving conditions are associated with low, medium, and high stress levels, respectively. 

\subsection{Modelling experiments}
\begin{table*} [h]
\caption{Results for each driver evaluated at test.}
	\centering
	\begin{tabular}{|l||c|c|c|c|c|c|c|c|c||c|}
		\hline
	     {\bf Method} & $D_1$ & $D_2$ & $D_3$ & $D_4$ & $D_5$ & $D_7$ & $D_9$ & $D_{10}$ & $D_{11}$ & {\bf Avg.} \\
		\hline 
		\hline
		Object presence -- Random Forest & 0.51  & 0.57 & 0.7 & 0.68  & 0.71  & 0.71 & 0.61 & 0.66 & 0.64 & 0.64\\
        \hline
        Object presence -- Linear SVM & 0.52 & 0.54 & 0.65 & 0.57 & 0.61 & 0.6 & 0.58 & 0.63 & 0.61 & 0.59\\
        \hline
		Object presence -- RBF-SVM & 0.48 & 0.51 & 0.65 & 0.56 & 0.65 & 0.61 & 0.57 & 0.62 & 0.62 & 0.58\\
		\hline
		\hline
		Single frame -- CNN & 0.56 & 0.62 & 0.72 & 0.71 & \bf{0.73} & 0.73 & 0.64 & \bf{0.74} & 0.70 & 0.68\\
		\hline
		\hline
		Video sequence -- TSN & \bf{0.6} & \bf{0.66}  & \bf{0.78} & \bf{0.87} & 0.72 & \bf{0.78} & \bf{0.68} & 0.71 &  \bf{0.72} & \bf{0.72} \\
		\hline
	\end{tabular}
	\label{tab:datasets_acc}
\end{table*}

\subsubsection{Image classification with object presence features}
\label{sec:class_feature_vect}

Our first modelling experiments consist of using classical machine learning methods on image features that encode the presence of objects in the image, as described in Sect. \ref{sec:meth_object_presence_features}. Tab. \ref{tab:datasets_acc} shows the accuracy obtained in each of the data splits and the corresponding average (see rows 1, 2 and 3). We observe that the accuracy obtained is significantly above chance (which would be $0.33$). This effort shows how a simple representation of the visual information can give relevant insights about driver self-reported stress. Notice that these results are supported by the observations reported in Sect. \ref{sec:stress_scene_obj}, suggesting that the appearance of particular objects in the driving scene may be a visual signature associated with stress.

\subsubsection{End-to-end image classification}


The results obtained by the end-to-end image CNN approach described in Sect.\ref{sec:methodology_image} are shown in Tab. \ref{tab:datasets_acc}, row 4. We observe, as expected, that the image CNN approach outperforms the classical ML methods tested before. 
Interestingly, we can see that the accuracy varies across the different drivers. The lowest accuracies are obtained for the $D_1$, $D_2$, and $D_9$ data splits. This can be explained by the fact that some drivers' perceived stress is significantly different from that of others.

\begin{figure}[h]
\centerline{\includegraphics[width=0.5\textwidth]{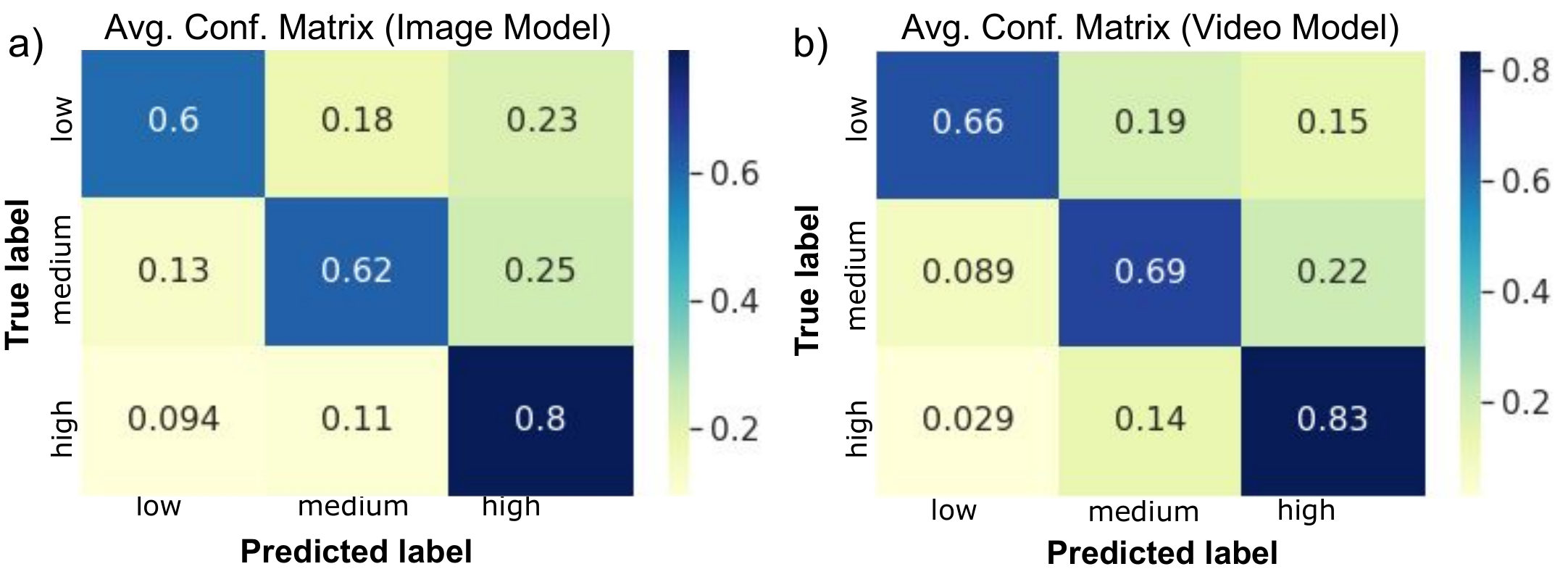}}
\caption{Average confusion matrices computed on (a) the test sets for the Image CNN Model and (b) the video TSN Model. }
\label{fig:conf_matrices}
\end{figure}


\begin{figure}[h]
\centerline{\includegraphics[width=0.5\textwidth]{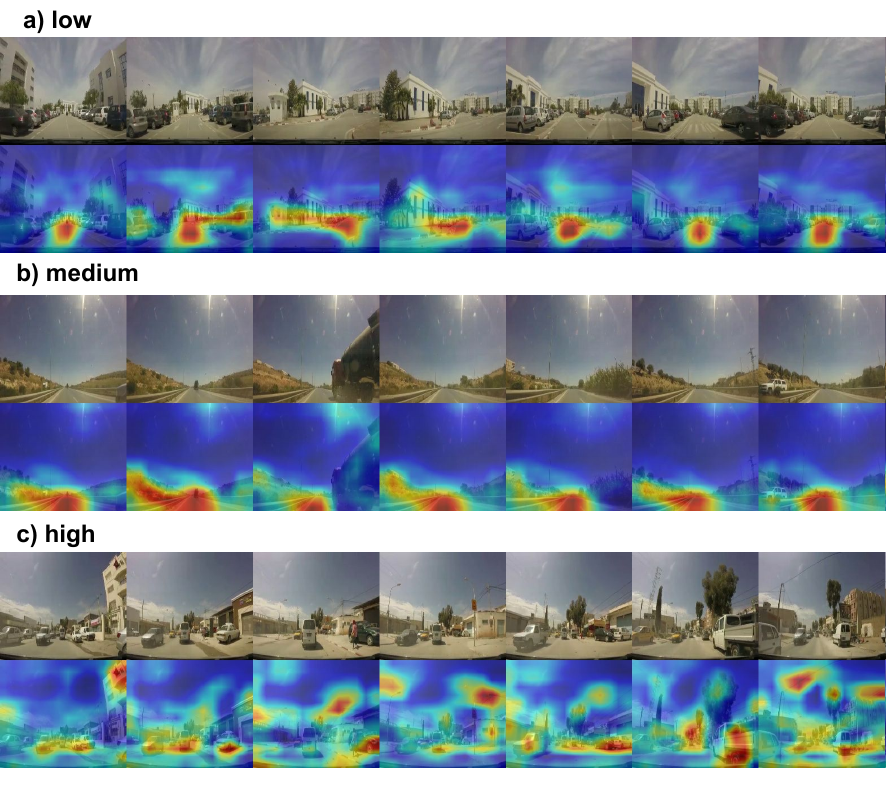}}
\caption{Class Activation Maps for two random examples of video sequences for low (a), medium (b), and high (c) stress measure classes.}
\label{fig:examples_cam}
\end{figure}

Notice that $D_1$ and $D_2$ involve test data related to two drivers Drv1 and Drv2, respectively. Drv1 repeated the experiment three times and Drv2 repeated the experiment twice. While the annotation by the observer was done in real-time, the validation and correction of the stress metric by the driver was done on different days. The validation session of the first drive could have happened after the second drive for both of these drivers, which might affect their recollection of their stress levels. $D_9$ considers the data of the participant Drv7 for the testing, and we learned that Drv7 lived in a foreign country for years and may have had a culturally different perception of the rated stress\footnote{We consulted the authors of the database for these details.}. More data are needed to make general conclusions about the possible impact of these factors.

Figure \ref{fig:conf_matrices} (a) shows the average confusion matrix for the image CNN model. Examples in the low-stress class that are wrongly classified are confused with medium and high in an even manner. On the contrary, we see that wrongly classified examples in the medium and high classes are rarely confused with the low-stress class. Wrongly classified medium-stress examples are usually classified as high, while wrongly classified high-stress examples are usually classified as medium. Our explanation is that medium-stress and high-stress examples are visually more similar than low-stress examples.

\subsubsection{End-to-end video classification}


We trained different TSN models (see Sect.\ref{sec:methodology_video}) by considering different video sequences of length $n=1$ seconds to $n=32$ seconds and $K=8$. Results show increments on the accuracy up to a time window of $n=20$ seconds, where we obtain the maximum mean accuracy of $0.72$. After that point, accuracy steadily decreases as we keep increasing $n$. Our thinking is that windows too short may not capture enough temporal contextual information, too large may provide too sparse of temporal contextual information while windows, while 20 seconds may be "just right" for discriminating driving events corresponding to varied levels of driver stress. 

Fig. \ref{fig:conf_matrices}. (b) shows the average confusion matrix computed across all of the test sets. Results and observations are similar to the ones obtained without considering temporal information: The highest confusion happens between medium and high classes. As we would expect, since the video model contains also temporal data, the correct classifications are higher than the ones obtained with the image CNN model.

Finally, Fig. \ref{fig:examples_cam} shows CAM visualizations for the video model. Common to all examples, it is interesting to see the general tendency to focus on the central part of the image, the region where the road is located. This fact is of special interest since it is also central for human driving tasks \cite{palazzi2018predicting} which may explain the fixation of the CNN on objects in that region. Now, comparing the CNN fixation maps obtained for different stress levels, we highlight the increasing complexity of the CAM for images with higher levels of stress. As elaborated in the introduction, increasing complexity within the driving task usually results in higher levels of stress. Our Class Activation Mappings also seem to indicate that the CNN requires more complex analysis to classify a scene as highly stressful. This fact is also compatible with recent findings relating urban safety with scene disorder \cite{bustos2021explainable}. 



\section{Conclusions}

In this paper, we hypothesized that drivers' subjective stress levels might be estimated directly from automated visual analysis of the driving scene. 
Using the driving scene videos and the corresponding human-driver-provided stress measures of the AffectiveROAD dataset, we trained three different Machine Learning approaches to estimate driver stress from the driving scenes. 
The best average accuracy across testing sets was obtained using a video CNN, beating a random model of $0.33$ and obtaining $0.72$ on leave-one-person out test data.
While more work is needed to test on larger and more diverse data sets, 
representing the variety of cultural and individual differences, these results suggest that the automated analysis of environmental context may contribute significantly to inferring driver affective states during real-world driving conditions.





\section*{Acknowledgment}

This work was partially supported by the Spanish Ministry of Science, Innovation and Universities, TIN2015-66951-C2-545-2-R and RTI2018-095232-B-C22. CB is supported by a PhD grant from the Universitat Oberta de Catalunya (UOC). We thank the CEA-LinkLab for providing us with the videos and details on the cohort involved in the study.  


\newpage
\bibliographystyle{IEEEtran}
\bibliography{IEEEabrv,references}

\begin{thebibliography}{10}
\providecommand{\url}[1]{#1}
\csname url@samestyle\endcsname
\providecommand{\newblock}{\relax}
\providecommand{\bibinfo}[2]{#2}
\providecommand{\BIBentrySTDinterwordspacing}{\spaceskip=0pt\relax}
\providecommand{\BIBentryALTinterwordstretchfactor}{4}
\providecommand{\BIBentryALTinterwordspacing}{\spaceskip=\fontdimen2\font plus
\BIBentryALTinterwordstretchfactor\fontdimen3\font minus
  \fontdimen4\font\relax}
\providecommand{\BIBforeignlanguage}[2]{{%
\expandafter\ifx\csname l@#1\endcsname\relax
\typeout{** WARNING: IEEEtran.bst: No hyphenation pattern has been}%
\typeout{** loaded for the language `#1'. Using the pattern for}%
\typeout{** the default language instead.}%
\else
\language=\csname l@#1\endcsname
\fi
#2}}
\providecommand{\BIBdecl}{\relax}
\BIBdecl

\bibitem{Lyu2017}
N.~Lyu, L.~Xie, C.~Wu, Q.~Fu, and C.~Deng, ``Driver’s cognitive workload and
  driving performance under traffic sign information exposure in complex
  environments: A case study of the highways in china,'' \emph{Int. journal of
  environmental research and public health}, vol.~14, no.~2, p. 203, 2017.

\bibitem{laumann2003selective}
K.~Laumann, T.~G{\"a}rling, and K.~M. Stormark, ``Selective attention and heart
  rate responses to natural and urban environments,'' \emph{Journal of
  environmental psychology}, vol.~23, no.~2, pp. 125--134, 2003.

\bibitem{mehler2009impact}
B.~Mehler, B.~Reimer, J.~F. Coughlin, and J.~A. Dusek, ``Impact of incremental
  increases in cognitive workload on physiological arousal and performance in
  young adult drivers,'' \emph{Transportation Research Record}, vol. 2138,
  no.~1, pp. 6--12, 2009.

\bibitem{jeon2014effects}
M.~Jeon, B.~N. Walker, and J.-B. Yim, ``Effects of specific emotions on
  subjective judgment, driving performance, and perceived workload,''
  \emph{Transportation research part F: traffic psychology and behaviour},
  vol.~24, pp. 197--209, 2014.

\bibitem{hennessy1999traffic}
D.~A. Hennessy and D.~L. Wiesenthal, ``Traffic congestion, driver stress, and
  driver aggression,'' \emph{Aggressive Behavior: Official Journal of the
  International Society for Research on Aggression}, vol.~25, no.~6, pp.
  409--423, 1999.

\bibitem{Jeon2016}
M.~Jeon, ``Don’t cry while you’re driving: sad driving is as bad as angry
  driving,'' \emph{Int. Journal of Human--Computer Interaction}, vol.~32,
  no.~10, pp. 777--790, 2016.

\bibitem{Magana2020}
V.~C. Maga{\~n}a, W.~D. Scherz, R.~Seepold, N.~M. Madrid, X.~G. Pa{\~n}eda, and
  R.~Garcia, ``The effects of the driver’s mental state and passenger
  compartment conditions on driving performance and driving stress,''
  \emph{Sensors}, vol.~20, no.~18, p. 5274, 2020.

\bibitem{hennessy1997relationship}
D.~A. Hennessy and D.~L. Wiesenthal, ``The relationship between traffic
  congestion, driver stress and direct versus indirect coping behaviours,''
  \emph{Ergonomics}, vol.~40, no.~3, pp. 348--361, 1997.

\bibitem{hill2007driver}
J.~D. Hill and L.~N. Boyle, ``Driver stress as influenced by driving maneuvers
  and roadway conditions,'' \emph{Transportation Research Part F: Traffic
  Psychology and Behaviour}, vol.~10, no.~3, pp. 177--186, 2007.

\bibitem{Bitkina2019}
O.~V. Bitkina, J.~Kim, J.~Park, J.~Park, and H.~K. Kim, ``Identifying traffic
  context using driving stress: A longitudinal preliminary case study,''
  \emph{Sensors}, vol.~19, no.~9, p. 2152, 2019.

\bibitem{chung2019methods}
W.-Y. Chung, T.-W. Chong, and B.-G. Lee, ``Methods to detect and reduce driver
  stress: a review,'' \emph{Int. Journal of Automotive Technology}, vol.~20,
  no.~5, pp. 1051--1063, 2019.

\bibitem{Healey2005}
J.~A. Healey and R.~W. Picard, ``Detecting stress during real-world driving
  tasks using physiological sensors,'' \emph{IEEE Trans. on Intelligent
  Transportation Systems}, vol.~6, no.~2, pp. 156--166, 2005.

\bibitem{Liu2018}
Y.~Liu and S.~Du, ``Psychological stress level detection based on electrodermal
  activity,'' \emph{Behavioural brain research}, vol. 341, pp. 50--53, 2018.

\bibitem{ElHaouij2018}
N.~Elhaouij, J.-M. Poggi, S.~Sevestre-Ghalila, R.~Ghozi, and M.~Ja{\"\i}dane,
  ``Affective{{ROAD}} system and database to assess driver's attention,'' in
  \emph{Proc. of the 33rd Annual ACM Symposium on Applied Computing}, 2018, pp.
  800--803.

\bibitem{ElhaouijPhDThesis}
N.~El~Haouij, ``Biosignals for driver's stress level assessment: functional
  variable selection and fractal characterization,'' Ph.D. dissertation,
  Universit{\'e} Paris-Saclay (ComUE), 2018.

\bibitem{WebsiteAffectiveRoad}
``Affective{{ROAD}} data (2018),''
  \url{https://www.media.mit.edu/groups/affective-computing/data/}, accessed:
  2021-04-03.

\bibitem{selvaraju2017grad}
R.~R. Selvaraju, M.~Cogswell, A.~Das, R.~Vedantam, D.~Parikh, and D.~Batra,
  ``Grad-cam: Visual explanations from deep networks via gradient-based
  localization,'' in \emph{Proc. of the IEEE Int. Conf. on Computer Vision},
  2017, pp. 618--626.

\bibitem{ZepfSurvey}
S.~Zepf, J.~Hernandez, A.~Schmitt, W.~Minker, and R.~W. Picard, ``Driver
  emotion recognition for intelligent vehicles: A survey,'' \emph{ACM Comput.
  Surv.}, vol.~53, no.~3, Jun. 2020.

\bibitem{Simon1996}
F.~Simon and C.~Corbett, ``Road traffic offending, stress, age, and accident
  history among male and female drivers,'' \emph{Ergonomics}, vol.~39, no.~5,
  pp. 757--780, 1996.

\bibitem{Bowen2020}
L.~Bowen, S.~L. Budden, and A.~P. Smith, ``Factors underpinning unsafe driving:
  A systematic literature review of car drivers,'' \emph{Transportation
  Research Part F: Traffic Psychology and Behaviour}, vol.~72, pp. 184--210,
  2020.

\bibitem{Nass2005}
C.~Nass, I.-M. Jonsson, H.~Harris, B.~Reaves, J.~Endo, S.~Brave, and
  L.~Takayama, ``Improving automotive safety by pairing driver emotion and car
  voice emotion,'' in \emph{CHI'05 extended abstracts on Human factors in
  computing systems}, 2005, pp. 1973--1976.

\bibitem{Hernandez2014}
J.~Hernandez, D.~McDuff, X.~Benavides, J.~Amores, P.~Maes, and R.~Picard,
  ``Autoemotive: bringing empathy to the driving experience to manage stress,''
  in \emph{Proc. of the 2014 Companion Publication on Designing Interactive
  Systems}, 2014, pp. 53--56.

\bibitem{Paredes2018}
P.~E. Paredes, Y.~Zhou, N.~A.-H. Hamdan, S.~Balters, E.~Murnane, W.~Ju, and
  J.~A. Landay, ``Just breathe: In-car interventions for guided slow
  breathing,'' \emph{Proc. of the ACM on Interactive, Mobile, Wearable and
  Ubiquitous Technologies}, vol.~2, no.~1, pp. 1--23, 2018.

\bibitem{Braun2019}
M.~Braun, J.~Schubert, B.~Pfleging, and F.~Alt, ``Improving driver emotions
  with affective strategies,'' \emph{Multimodal Technologies and Interaction},
  vol.~3, no.~1, p.~21, 2019.

\bibitem{Rastgoo2018}
M.~N. Rastgoo, B.~Nakisa, A.~Rakotonirainy, V.~Chandran, and D.~Tjondronegoro,
  ``A critical review of proactive detection of driver stress levels based on
  multimodal measurements,'' \emph{ACM Computing Surveys (CSUR)}, vol.~51,
  no.~5, pp. 1--35, 2018.

\bibitem{Nvemcova2020}
A.~N{\v{e}}mcov{\'a}, V.~Svozilov{\'a}, K.~Bucsuh{\'a}zy, R.~Sm{\'\i}{\v{s}}ek,
  M.~M{\'e}zl, B.~Hesko, M.~Bel{\'a}k, M.~Bil{\'\i}k, P.~Maxera, M.~Seitl
  \emph{et~al.}, ``Multimodal features for detection of driver stress and
  fatigue,'' \emph{IEEE Trans. on Intelligent Transportation Systems}, 2020.

\bibitem{Rimini2001}
M.~Rimini-Doering, D.~Manstetten, T.~Altmueller, U.~Ladstaetter, and M.~Mahler,
  ``Monitoring driver drowsiness and stress in a driving simulator,'' 2001.

\bibitem{bustos2021explainable}
C.~Bustos, D.~Rhoads, A.~Sol{\'e}-Ribalta, D.~Masip, A.~Arenas, A.~Lapedriza,
  and J.~Borge-Holthoefer, ``Explainable, automated urban interventions to
  improve pedestrian and vehicle safety,'' \emph{Transportation Research Part
  C: Emerging Technologies}, vol. 125, p. 103018, 2021.

\bibitem{Ren2016}
S.~Ren, K.~He, R.~Girshick, and J.~Sun, ``Faster r-cnn: towards real-time
  object detection with region proposal networks,'' \emph{IEEE Trans. on
  Pattern Analysis and Machine Intelligence}, vol.~39, no.~6, pp. 1137--1149,
  2016.

\bibitem{Zhao2017}
H.~Zhao, J.~Shi, X.~Qi, X.~Wang, and J.~Jia, ``Pyramid scene parsing network,''
  in \emph{Proc. of the IEEE Conf. on Computer Vision and Pattern Recognition},
  2017, pp. 2881--2890.

\bibitem{Geiger2012}
A.~Geiger, P.~Lenz, and R.~Urtasun, ``Are we ready for autonomous driving? the
  kitti vision benchmark suite,'' in \emph{2012 IEEE Conf. on Computer Vision
  and Pattern Recognition}.\hskip 1em plus 0.5em minus 0.4em\relax IEEE, 2012,
  pp. 3354--3361.

\bibitem{zhou2017places}
B.~Zhou, A.~Lapedriza, A.~Khosla, A.~Oliva, and A.~Torralba, ``Places: A 10
  million image database for scene recognition,'' \emph{IEEE trans. on pattern
  analysis and machine intelligence}, vol.~40, no.~6, pp. 1452--1464, 2017.

\bibitem{Ding2020}
L.~Ding, M.~Glazer, M.~Wang, B.~Mehler, B.~Reimer, and L.~Fridman, ``Mit-avt
  clustered driving scene dataset: Evaluating perception systems in real-world
  naturalistic driving scenarios,'' in \emph{2020 IEEE Intelligent Vehicles
  Symposium (IV)}.\hskip 1em plus 0.5em minus 0.4em\relax IEEE, pp. 232--237.

\bibitem{Kato2011}
T.~Kato, H.~Kawanaka, M.~S. Bhuiyan, and K.~Oguri, ``Classification of positive
  and negative emotion evoked by traffic jam based on electrocardiogram (ecg)
  and pulse wave,'' in \emph{14th Int. IEEE Conf. on Intelligent Transportation
  Systems (ITSC)}, 2011, pp. 1217--1222.

\bibitem{Ihme2018}
K.~Ihme, C.~D{\"o}meland, M.~Freese, and M.~Jipp, ``Frustration in the face of
  the driver: a simulator study on facial muscle activity during frustrated
  driving,'' \emph{Interaction Studies}, vol.~19, no.~3, pp. 487--498, 2018.

\bibitem{bulo2018place}
S.~R. Bulo, L.~Porzi, and P.~Kontschieder, ``In-place activated batchnorm for
  memory-optimized training of dnns,'' in \emph{Proc. of the IEEE Conf. on
  Computer Vision and Pattern Recognition}, 2018, pp. 5639--5647.

\bibitem{neuhold2017mapillary}
G.~Neuhold, T.~Ollmann, S.~Rota~Bulo, and P.~Kontschieder, ``The mapillary
  vistas dataset for semantic understanding of street scenes,'' in
  \emph{Proceedings of the IEEE Int. Conf. on Computer Vision}, 2017, pp.
  4990--4999.

\bibitem{simonyan2014very}
K.~Simonyan and A.~Zisserman, ``Very deep convolutional networks for
  large-scale image recognition,'' \emph{arXiv preprint arXiv:1409.1556}, 2014.

\bibitem{wang2016temporal}
L.~Wang, Y.~Xiong, Z.~Wang, Y.~Qiao, D.~Lin, X.~Tang, and L.~Van~Gool,
  ``Temporal segment networks: Towards good practices for deep action
  recognition,'' in \emph{European Conf. on Computer Vision}.\hskip 1em plus
  0.5em minus 0.4em\relax Springer, 2016, pp. 20--36.

\bibitem{zhou2016learning}
B.~Zhou, A.~Khosla, A.~Lapedriza, A.~Oliva, and A.~Torralba, ``Learning deep
  features for discriminative localization,'' in \emph{Proc. of the IEEE Conf.
  on Computer Vision and Pattern Recognition}, 2016, pp. 2921--2929.

\bibitem{palazzi2018predicting}
A.~Palazzi, D.~Abati, F.~Solera, R.~Cucchiara \emph{et~al.}, ``Predicting the
  driver's focus of attention: the dr (eye) ve project,'' \emph{IEEE Trans. on
  Pattern Analysis and Machine Intelligence}, vol.~41, no.~7, pp. 1720--1733,
  2018.

\end{thebibliography}

\end{document}